\documentclass[a4paper, 10pt, conference]{ieeeconf}      
\usepackage{FG2026}
\usepackage{amsmath, amssymb}
 \usepackage{graphicx}
 \usepackage{subcaption}
 \usepackage{xcolor}
\FGfinalcopy 

\IEEEoverridecommandlockouts                              
\def\FGPaperID{103} 

\title{\LARGE \bfseries\textbf{
 3D Temporal Analysis for Autism Spectrum Disorder Screening During Attention Tasks}
}

\author{\parbox{16cm}{\centering
    {\large Inam Qadir, Elizabeth B Varghese, Dena Al-Thani, Marwa Qaraqe}\\
    {\normalsize
    College of Science and Engineering, 
Hamad Bin Khalifa University, Qatar Foundation, Doha, Qatar\\}}
    \thanks{The research reported in this publication was supported by the Qatar Research Development and Innovation Council (ARG01-0508-230097).}
}

\usepackage{fancyhdr}

\begin{document}

\maketitle
\thispagestyle{fancy}

\begin{abstract}

Accurate Autism Spectrum Disorder (ASD) screening for school-age children is crucial to identify cases that may have been missed earlier and to enable timely interventions supporting social, cognitive, and academic development. Current ASD screening relies on subjective assessments and 2D analysis methods that fail to capture spatial displacement patterns characteristic of ASD behaviors. In this study, a novel 3D temporal analysis framework is presented, built on top of DECA (Detailed Expression Capture and Animation), a 3D modeling framework, to extract comprehensive head pose parameters (including translational components $T_x, T_y, T_z$) and facial expressions independent of pose variations. LSTM and GRU-based temporal classifiers were trained on the extracted 3D features from video data collected from 39 participants (19 ASD, 20 TD) aged 7-12 years during Virtual Reality-Continuous Performance Test tasks. 
The GRU-based models demonstrated superior performance, with 3D head pose features achieving 83.9\% accuracy and 3D facial features reaching 81.4\% accuracy, outperforming 2D baseline approaches by 10.7\% and 7.5\%, respectively. Furthermore, multimodal fusion of 3D head pose and facial features with PCA-based dimensionality reduction achieved the highest accuracy of 84.6\%, outperforming unimodal approaches. This work establishes a foundation for objective, automated screening tools addressing current diagnostic limitations in ASD identification for school-age populations.

\end{abstract}

\section{INTRODUCTION}
Autism Spectrum Disorder (ASD) is a complex neurodevelopmental condition characterized by challenges in social interaction, communication difficulties, and repetitive behavioral patterns \cite{wing1979severe}. With  prevalence rate reaching 1 in 31 children in the united states according to the Centers for Disease Control and Prevention \cite{maenner2023prevalence}, this substantial rate underscores the critical importance of effective screening mechanisms. These prevalence estimates were mainly based on the standard diagnostic measures such as Autism Diagnostic Observation Schedule (ADOS) \cite{kamp2018diagnostic}, Autism Diagnostic Interview-Revised (ADI-R) \cite{lord1994autism}, Childhood Autism Rating Scale (CARS) \cite{schopler2010childhood}, and Modified Checklist for Autism in Toddlers (M-CHAT) \cite{robins2001modified}. \\
\begin{figure}[htb!]
    \centering
    \includegraphics[width=0.40\textwidth]{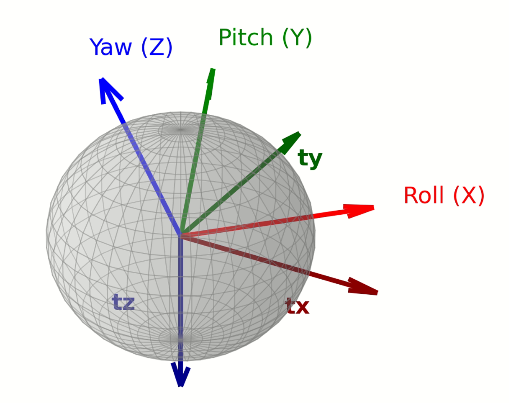}
    \caption{Illustration of 3D head pose parameters showing rotational and translational components.}
    \label{fig:component}
\end{figure}
The current assessment methods are inherently subjective and dependent on clinician expertise, requiring time-intensive evaluations (40-60 minutes for ADOS, 2-3 hours for ADI-R) and specialized multidisciplinary teams. Such limitations often result in delayed diagnosis, limited access to services, and assessment discrepancies across healthcare providers \cite{li2022automatic}, emphasizing the need for more objective, automated screening approaches. While early diagnosis is ideal, some cases—particularly those with subtle symptoms or late symptom onset may be missed in preschool years. This highlights the critical need for reliable and timely screening for school-age children, ensuring that previously undiagnosed cases are identified and provided with appropriate interventions to support social, cognitive, and academic development \cite{barry2016school}.

In addition, the phenomenon of \textit{masking}, particularly prevalent in females and individuals with milder ASD symptoms, often leads to delayed or missed diagnoses as individuals conceal their difficulties through learned social behaviors \cite{lockwood2021barriers}. Although screening serves as a preliminary identification step during routine pediatric visits or educational settings, formal diagnosis requires extensive evaluation by specialists such as pediatric neurologists and developmental pediatricians using specialized assessment methods and clinical settings \cite{filipek2013screening}. However, these diagnostic procedures are impractical for widespread screening initiatives due to their resource-intensive nature \cite{song2022early}. This limitation, combined with current reliance on time-intensive behavioral assessments, creates a significant gap in effective screening approaches for school-age children, a challenge that this study aims to address through objective 3D temporal analysis using advanced computational methods.\\

Recently, advances in computer vision approaches have explored appearance-based features such as head pose and facial expression analysis \cite{gokmen2024detecting,song2022early} for ASD screening. Even though these methods demonstrate potential, they remain constrained by fundamental 2D representation limitations. Specifically, traditional head pose estimation methods extract only Euler angles (yaw, pitch, roll) from image projections, failing to capture translational movements ($T_x, T_y, T_z$) that reveal spatial displacement patterns, approach-avoidance behaviors, and repetitive movements characteristic of self-stimulatory behaviors in ASD \cite{leekam2011restricted}. Concurrently, 2D facial expression analysis methods struggle to decouple expressions from head pose and individual morphological differences, leading to  feature representations confounded by factors such as lighting conditions, camera perspectives, and identity-related facial structure variations \cite{sariyanidi2025beyond}. 
These constraints reduce the discriminative power of behavioral markers and may contribute to inconsistent screening performance across populations and recording conditions.

To address these limitations, this study presents a 3D temporal analysis framework built on top of DECA (Detailed Expression Capture and Animation) \cite{DECA:Siggraph2021} to extract robust 3D head pose and facial expression features. The framework employs Long short-term memory (LSTM) \cite{graves2012long} and Gated recurrent unit (GRU) \cite{chung2014empirical} based temporal modeling to capture sequential behavioral patterns. The methodology is validated using a specialized video dataset collected during Virtual Reality-Continuous Performance Test (VR-CPT) attention assessment tasks. This setup is motivated by research demonstrating that attention processing presents distinctive patterns in children with ASD \cite{johnson2004attention,varghese2024attention}, where both facial expressions and head movements serve as valuable indicators of cognitive engagement \cite{asteriadis2011importance,banire2021face}. Through separate and multimodal analysis of 3D head pose features incorporating rotational and translational components alongside 3D facial expression features independent of pose and identity variations, the framework provides enhanced behavioral markers applicable across various assessment contexts, including structured attention tasks, free play scenarios, and naturalistic social interactions. The main contributions of this work include:\\
\begin{itemize}
\item Introduced the use of 3D head pose translation vector components \((T_x, Ty, T_z)\) to capture spatial displacement patterns, demonstrating superior discriminative power for ASD screening compared to 2D representations.

\item Developed a 3D feature extraction pipeline built on top of DECA for robust and pose-invariant head pose and facial expression analysis.

\item Utilized attention-related behavioral video data collected during VR-CPT tasks in a virtual classroom environment to study ASD-specific behavioral patterns.

\item Designed and implemented LSTM- and GRU-based temporal classifiers to process sequential behavioral patterns, achieving improved classification performance with 3D features over traditional 2D approaches.

\end{itemize}
\section{Related Work}
ASD screening methodology has evolved from traditional observational approaches to emerging computational solutions. Conventional protocols rely on behavioral observations, caregiver interviews, and standardized assessments, with early frameworks like the Early Screening of Autistic Traits Questionnaire (ESAT) \cite{oosterling2010advancing} and parent-report instruments \cite{wieckowski2021early} primarily targeting early childhood. The critical need for school-age screening tools was addressed by Ehlers et al. \cite{ehlers1999screening} through the Autism Spectrum Screening Questionnaire (ASSQ) for children aged 7-16 years. Sheldrick et al. \cite{sheldrick2017age} further validated that children with subtle ASD manifestations often remain unidentified until school age. However, these established methodologies face significant limitations including assessment subjectivity \cite{brian2019standards}, extensive time requirements \cite{simeoli2024using}, and inconsistent clinical outcomes \cite{lord2018autism}. These challenges have motivated the exploration of objective, computational methods to provide quantifiable metrics for ASD screening in school-age populations.

With advancements in computational capabilities and deep learning technologies\cite{qadir2025fusion}, ASD screening research has evolved towards facial image analysis approaches. Elangovan et al. \cite{elangovan2024fusion} proposed the Fusion of Transfer Learning with the Dandelion Algorithm, demonstrating classification performance through integration of MobileNetV2, DenseNet201, and ResNet50 models. Lee et al. \cite{lu2021deep} addressed diagnostic disparities by developing race-specific models using VGG16 architecture, while Kanwal et al. \cite{kanwal2024hybrid} introduced a framework integrating ConvNeXt-T models with embedding clusters. Despite these advances, 2D facial analysis methods cannot decouple expressions from head pose variations and morphological differences, leading to confounded feature representations \cite{sariyanidi2025beyond}.   Specifically, these limitations manifest as head rotation being misclassified as eyebrow movements or mouth shape changes, lighting direction variations causing identical neutral expressions to be classified as different emotional states, and facial structure differences between individuals causing the same expression intensity to produce varying feature measurements across subjects \cite{canedo2019facial}.

Parallel research has explored head pose analysis for ASD screening, demonstrating measurable movement differences between populations. Martin et al. \cite{martin2018objective} found children with ASD exhibited greater yaw displacement and higher movement velocity during social stimuli, while Dawson et al. \cite{dawson2018atypical} showed toddlers with ASD demonstrated head movement rates 1.53-2.45 times higher than controls. Gokmen et al. \cite{gokmen2024detecting} applied kinesics theory to classify head movement patterns, achieving 79.7\% and 76.7\% accuracy in adolescents and young adults, respectively. However, traditional head pose estimation methods extract only Euler angles (yaw, pitch, roll), overlooking translational components that could provide additional behavioral insights. By incorporating translation vectors (Tx, Ty, Tz) from the DECA model, this study introduces a significant technical advancement: these parameters capture three-dimensional spatial displacement of the head, enabling detection of approach-avoidance tendencies, subtle positional shifts, and repetitive movement trajectories characteristic of ASD motor behaviors \cite{leekam2011restricted}. 
This richer representation provides novel behavioral markers that extend beyond rotation-only analysis, facilitating comprehensive behavioral assessment through temporal modeling of 3D geometric representations for enhanced ASD screening performance.


\section{Methodology}

The proposed methodology for screening ASD and TD (Typical Developing) processes video sequences captured during attention tasks to analyze temporal facial dynamics and head pose variations. The approach comprises two main components: (1) 3D feature extraction built on top of the DECA framework\cite{DECA:Siggraph2021} to derive robust facial feature vectors and head pose parameters from video frames, and (2) temporal modeling of Recurrent Neural Networks (RNNs) to model sequential behavioral patterns.

\subsection{Feature Extraction}
\label{subsec:FE}
\begin{figure*}[!t]
    \centering
    \includegraphics[width=0.8\textwidth]{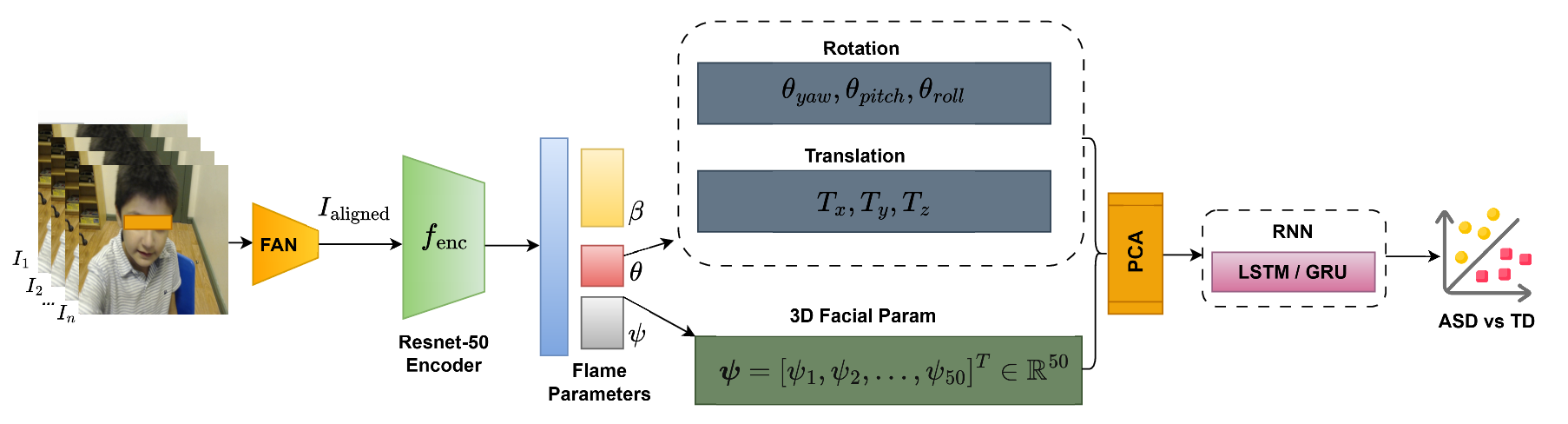}
    \caption{Overview of the feature extraction component. Video frames undergo preprocessing and encoding via ResNet-50 to extract FLAME model parameters, specifically 50-dimensional facial expression features and 6-dimensional head pose parameters (yaw, pitch, roll, and 3D translation), which are then fed to RNN-based temporal classifiers for ASD screening.}
    \label{fig:framework_diagram}
\end{figure*}

The feature extraction component is built on top of DECA, a framework trained on in-the-wild images without paired 3D supervision. DECA regresses 3D facial feature vectors capturing expression-dependent deformations and 3D head pose parameters from each input. The extraction pipeline consists of four stages: image preprocessing, latent encoding, FLAME(Faces Learned with an Articulated Model and Expressions) parameter prediction  where the latent features are decoded into FLAME's parametric representation that separates facial identity, expression, and pose components\cite{li2017learning}, and targeted feature selection, as depicted in the Figure. \ref{fig:framework_diagram}.

\subsubsection{Image Preprocessing}
Each video frame \( I \in \mathbb{R}^{H \times W \times 3} \) undergoes preprocessing to isolate and normalize the facial region. A face-alignment network, such as FAN\cite{bulat2017far}, detects 68 2D facial landmarks, which inform cropping and affine transformations to yield a standardized aligned frame \( I_{\text{aligned}} \in \mathbb{R}^{224 \times 224 \times 3} \). This alignment mitigates extraneous variations in scale, orientation, and position, ensuring reliable input for the subsequent encoding step.

\subsubsection{Latent Feature Encoding}
The framework's robustness against variations in illumination, occlusions, and poses prevalent in children's videos is attributed to its ResNet-50 encoder. The aligned frame \( I_{\text{aligned}} \) is processed by ResNet-50\cite{he2016deep} encoder \( f_{\text{enc}} \), pretrained on VGGFace2 for identity recognition, to produce a latent feature vector \( z \in \mathbb{R}^{236} \)
\begin{equation}
z = f_{\text{enc}}(I_{\text{aligned}}; \phi),
\end{equation}
where \( \phi \) represents the encoder's parameters. This compact representation captures essential spatial and appearance information, facilitating the regression of 3D parametric models in the following phase.

\subsubsection{FLAME Parameter Prediction}
The latent vector $\mathbf{z}$ contains estimated parameters of the FLAME 3D morphable model\cite{FLAME:SiggraphAsia2017}, which combines linear expression spaces with articulation for realistic face modeling
\begin{equation}
(\boldsymbol{\beta}, \boldsymbol{\psi}, \boldsymbol{\theta}, \boldsymbol{\alpha}, \mathbf{c}, \mathbf{l}) = \text{extract}(\mathbf{z}),
\end{equation}
where:
\begin{itemize}
\item $\boldsymbol{\psi} \in \mathbb{R}^{50}$: Expression parameters forming the 3D facial feature vector, capturing deformable geometry such as mouth and brow movements.
\item $\boldsymbol{\theta} \in \mathbb{R}^{6}$: Head pose parameters encompassing three Euler angles and three translation components:
    \begin{itemize}
    \item \textbf{Yaw} ($\theta_{yaw}$): Horizontal head rotation (left-right turning)
    \item \textbf{Pitch} ($\theta_{pitch}$): Vertical head rotation (up-down nodding)  
    \item \textbf{Roll} ($\theta_{roll}$): Axial head rotation (side-to-side tilting)
    \item \textbf{Translation} $(T_x, T_y, T_z)$: 3D spatial displacement
    \end{itemize}
\item $\boldsymbol{\beta}, \boldsymbol{\alpha}, \mathbf{c}, \mathbf{l}$: Shape, albedo, camera, and lighting parameters.
\end{itemize}


\subsubsection{Feature Configurations}
The analysis focuses on the expression parameters $\psi$ and head pose parameters $\theta$, which provide comprehensive 3D behavioral representations. Based on these, three configurations are evaluated: (a) 3D facial expressions ($\psi$), (b) 3D head pose ($\theta$), and (c) multimodal fusion concatenating both features,
\begin{equation}
    x_{\text{fused}} = [\psi; \theta] \in \mathbb{R}^{56}.
    \label{eq:fused}
\end{equation}

For multimodal fusion, expression and head pose features are synchronized at the frame level via shared video and frame identifiers, ensuring $\boldsymbol{\psi}^{(t)}$ and $\boldsymbol{\theta}^{(t)}$ correspond to identical temporal instants. To address the high dimensionality of the fused feature space relative to the limited sample size, Principal Component Analysis (PCA)\cite{mackiewicz1993principal} is applied for dimensionality reduction.The RNN implicitly learns cross-modal dependencies through its recurrent processing of the concatenated sequence.
For comparative evaluation, 2D facial features are also extracted using Py-Feat~\cite{cheong2023py}, yielding 20-dimensional action unit intensities as baseline features. Py-Feat is used as a representative 2D facial analysis baseline, as it provides OpenFace-compatible action unit features and 2D head pose descriptors commonly used in behavioral analysis. All features are temporally sequenced across video frames for subsequent RNN-based classification.


\subsection{Temporal Modeling}
The temporal modeling component processes the sequenced features extracted from the video frames to capture dynamic behavioral patterns for ASD screening. Given the sequential nature of facial expressions and head pose variations, RNN architectures are employed to model these temporal dependencies. Specifically, both LSTM and GRU variants are implemented to classify sequences as ASD or TD based on learned behavioral patterns over time.

\subsubsection{Input Preparation}

For a video consisting of $T$ frames, the input sequence is formed as 
$X = [x_{1}, x_{2}, \ldots, x_{T}]$, where each $x_{t}$ represents the feature vector at time $t$. The feature vectors correspond to the extracted facial expressions and head pose parameters as detailed in Section \ref{subsec:FE}. Sequences are zero-padded to the maximum length in the batch for efficient processing. These prepared sequences are then fed to RNN-based temporal classifiers for behavioral pattern analysis.
\subsubsection{RNN Framework}

\begin{figure}[!t]
    \centering
    \includegraphics[width=0.5\textwidth]{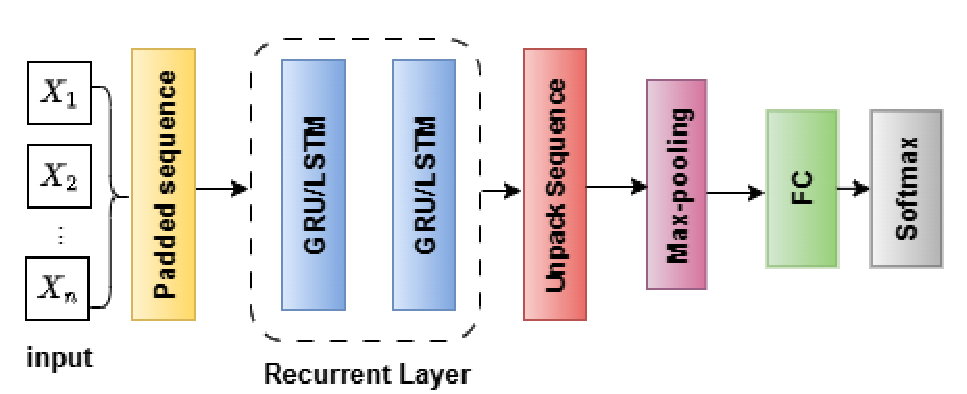}
    \caption{RNN-based temporal modeling architecture for sequential behavioral pattern classification..}
    \label{fig:model}
\end{figure}

A tailored RNN-based classifier is developed that can be configures with standard LSTM and GRU as the underlying recurrent cell type, as illustrated in Figure \ref{fig:model}. The core temporal modeling follows a general RNN formulation where the hidden state at time $t$ encodes information from the current input and previous temporal context

\begin{equation}
h_{t} = \text{RNN}(x_{t}, h_{t-1}; \gamma),
\end{equation}

\noindent where $h_{t}$ represents the hidden state vector at time $t$ that encapsulates the accumulated temporal information up to the current frame, $\gamma$ represents the learnable parameters. The classifier architecture remains consistent across both variants, with only the internal cell computations differing based on the selected recurrent unit.

For LSTM-based configurations, the classifier addresses long-term dependency modeling through a cell state mechanism controlled by three gates: input, forget, and output gates. These gates enable selective memory retention, valuable for capturing subtle behavioral patterns such as atypical facial expression timing and irregular head movement sequences across extended video sequences. The input gate controls which new facial or pose information to store, the forget gate determines which previous behavioral patterns to retain, and the output gate regulates how much accumulated behavioral context influences the current prediction.

Alternatively, when configured with GRU cells, the classifier provides a computationally efficient alternative with two gates: reset and update gates. The reset gate determines past information incorporation, while the update gate balances information retention and integration. This simplified gating mechanism often proves more effective for behavioral sequence modeling due to reduced parameter complexity, which can mitigate overfitting on relatively small and heterogeneous datasets typical of child behavioral studies.

\subsubsection{Sequence Aggregation Strategy}
To obtain fixed-size representations from variable-length sequences, temporal max-pooling is employed exclusively over the hidden states of the final recurrent layer

\[
h_{\text{final}} = \max_{t=1}^{T} h_t, \tag{8}
\]
where $h_t \in \mathbb{R}^{H \times D}$ represents the hidden state at timestep $t$. The pooled representation is then directly fed to the classification head through dropout regularization.

This aggregation strategy is selected over alternatives like mean-pooling or attention mechanisms because max-pooling captures the most salient behavioral moments,  such as sudden shifts in gaze, fleeting atypical expressions, or transient repetitive movements, which are often brief but highly discriminative in ASD screening contexts.
\subsubsection{Classification Module}
The aggregated representation undergoes dropout regularization followed by binary classification

\[
\hat{y} = \text{softmax}(W_{c} \cdot \text{Dropout}(h_{\text{final}}) + b_{c}), \tag{9}
\]

\noindent where $W_{c} \in \mathbb{R}^{2 \times H}$ and $b_{c} \in \mathbb{R}^{2}$ are learnable parameters, and $H$ represents the hidden dimension.

\section{Experiments}
This section presents a comprehensive evaluation of the proposed multimodal 3D temporal analysis framework for ASD screening. The effectiveness of the approach is assessed through systematic experiments comparing different feature modalities and temporal modeling architectures. The experimental design includes ablation studies to evaluate individual component contributions and comparative analysis against baseline approaches to validate the benefits of 3D temporal modeling. A specialized dataset collected during controlled attention assessment tasks is utilized for these evaluations.

\subsection{Dataset Description and Acquisition}
\label{subsec:dataset des}

\begin{figure}[!t]
  \centering
  \includegraphics[width=\columnwidth]{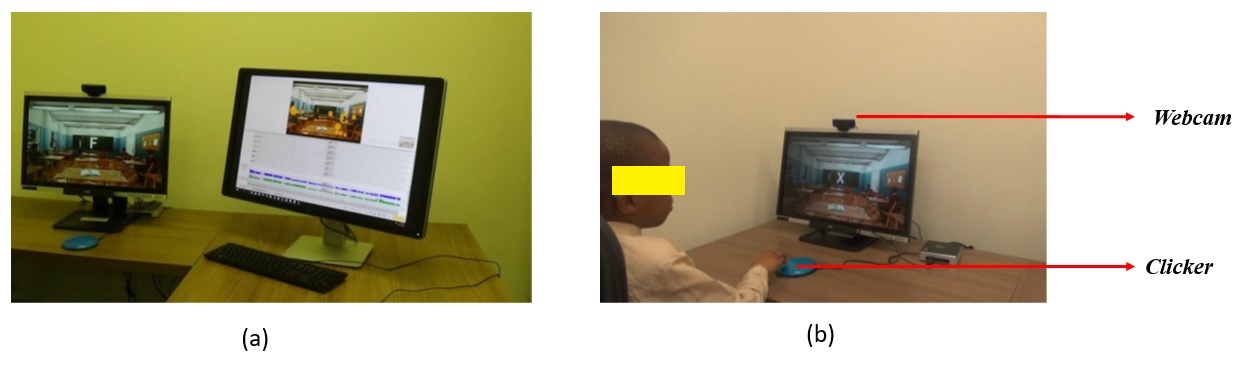}
  \caption{(a) Testing environment with two monitors- one for the participant and one for the researcher. (b) Sample virtual classroom with attached webcam and clicker}
  \label{fig:setup}
\end{figure}

The dataset used in this study was collected within a standardized Virtual Reality-Continuous Performance Test (VR-CPT) environment using a dual-monitor configuration: a 24-inch monitor for participants and a 34-inch monitor for researcher observation, as depicted in Figure \ref{fig:setup}. A webcam positioned on the participant's monitor recorded the children behavioral data.

The VR-CPT assessed selective and sustained attention through a comprehensive task where participants monitored letters displayed on a virtual blackboard, responding to target stimulus \texttt{X} while inhibiting responses to other letters. The task incorporated varying complexity levels through different presentation modes and environmental distractions.
The dataset comprises 39 participants (19 ASD, 20 TD) aged 7-12 years. All ASD participants were diagnosed by medical practitioners using DSM-IV-TR\cite{american2000quick} criteria, with inclusion limited to mild and moderate cases. The Childhood Autism Spectrum Test (CAST) was administered as supplementary validation, with ASD participants scoring above 15 and TD participants below this threshold. Detailed demographics are presented in Table \ref{tab:demographics}.

\begin{table}[h]
\centering
\caption{Demographic information showing mean values and standard deviations for age and CAST scores.}
\label{tab:demographics}
\begin{tabular}{l c c c}
\hline\noalign{\smallskip}
\textbf{Characteristics} & \textbf{ASD(n=19)} & \textbf{TD (n=20)} & \textbf{$p$ value} \\
\noalign{\smallskip}\hline\noalign{\smallskip}
Age(mean$\pm$std) & 8.57$\pm$1.40  & 8.58$\pm$1.38 & $p$ = 0.974\\
CAST Score & 17.75$\pm$2.04 & 5.7$\pm$3.2 & $p < $ 0.001\\
Gender(Male/Female) & 17/2 & 12/8 & N.A\\
\noalign{\smallskip}\hline
\end{tabular}
\end{table}

The VR-CPT task videos were recorded at 30 fps using a webcam positioned on the participant's monitor, with average duration of 3-4 minutes per participant, yielding up to 5,400-7,200 frames per video after quality filtering. Features were extracted frame-by-frame without temporal windowing, and each participant's complete frame sequence was processed as a single input to the RNN. Temporal max-pooling across all hidden states produced a fixed-size representation for classification, enabling the model to capture salient behavioral moments throughout the assessment.

The study received institutional review board approval, with data collection proceeding only after securing parental consent and participant assent. Data quality was ensured through real-time monitoring and post-collection screening protocols.

\subsection{Experimental Setup and Performance Metrics}

 Performance is evaluated using standard classification metrics: accuracy, precision, recall, and F1-score. 
 All experiments were executed in Python 3.10.13 using NVIDIA GeForce RTX 4090 GPUs with CUDA 12.3 for acceleration.

\subsection{Results and Discussion}


Following feature extraction as described in \ref{subsec:FE}, LSTM and GRU based RNNs are employed to classify ASD versus TD participants using temporal features. The optimal model architectures were determined through extensive hyperparameter tuning across multiple configurations, including RNN hidden state dimensions (64, 128, 256), number of recurrent layers (1-3), dropout regularization rates (0.1-0.3), learning rates (1e-4 to 1e-2), batch sizes (8, 16, 32), and bidirectional processing options. All models were trained for 100 epochs with early stopping applied when validation performance plateaued with 15 patience. The search space encompassed both architectural parameters (hidden size, layer depth, bidirectional configuration) and training hyperparameters (learning rate, batch size, dropout) to identify optimal configurations for each feature modality and RNN variant. The optimal hyperparameter and architectural configurations for 3D feature-based models are presented in Table \ref{tab:3d_hyperparameters}.

\begin{table}[h]
\centering
\caption{Optimal hyperparameters for 3D feature-based RNN architectures.}
\label{tab:3d_hyperparameters}
\resizebox{\columnwidth}{!}{
\begin{tabular}{l|cc|cc|cc}
\hline
\textbf{Parameter} & \multicolumn{2}{c|}{\textbf{3D Facial}} & 
\multicolumn{2}{c|}{\textbf{3D Head Pose}} & 
\multicolumn{2}{c}{\textbf{3D Fusion}} \\
 & \textbf{GRU} & \textbf{LSTM} & \textbf{GRU} & \textbf{LSTM} & \textbf{GRU} & \textbf{LSTM} \\
\hline
Hidden Size & 128 & 64 & 128 & 64 & 64 & 128 \\
Num Layers & 2 & 2 & 2 & 1 & 3 & 2 \\
Dropout & 0.2 & 0.2 & 0.1 & 0.2 & 0.3 & 0.1 \\
Learning Rate & 0.0001 & 0.0001 & 0.0001 & 0.0001 & 0.0001 & 0.0001 \\
Batch Size & 8 & 16 & 16 & 8 &16 & 32 \\
Bidirectional & No & Yes & No & No & Yes & No \\
\hline
\end{tabular}
}
\end{table}
The classification performance using 3D features extracted via the DECA framework is shown in Table \ref{tab:3d_results}. For unimodal features, 3D head pose features achieved the highest accuracy of 83.9\% using GRU architecture, while 3D facial features reached 81.4\% accuracy with GRU, consistently outperforming LSTM across both modalities. For multimodal fusion, PCA-based dimensionality reduction was applied to address the high feature-to-sample ratio of the concatenated 56-dimensional feature vector $\mathbf{x}_{\text{fused}}$ (Eqn. \ref{eq:fused}). The multimodal fusion achieved the highest overall performance, with LSTM reaching 84.6\% accuracy, outperforming the best unimodal results. The standard deviations across folds (0.065-0.165) suggest reasonable model stability despite the limited sample size.

\begin{table}[h]
\centering
\caption{Classification performance using 3D features with 5-fold cross-validation (mean $\pm$ std).}
\label{tab:3d_results}
\resizebox{\columnwidth}{!}{%
\begin{tabular}{l|cccc}
\hline
\textbf{Model} & \textbf{Accuracy} & \textbf{Precision} & \textbf{Recall} & \textbf{F1-Score} \\
\hline
\multicolumn{5}{c}{\textit{3D Facial Features}} \\
GRU  & $0.814\pm0.165$ & $0.833\pm0.211$ & $0.900\pm0.200$ & $0.827\pm0.150$ \\
LSTM & $0.786\pm0.099$ & $0.754\pm0.156$ & $0.950\pm0.100$ & $0.822\pm0.070$ \\
\hline
\multicolumn{5}{c}{\textit{3D Head Pose}} \\
GRU  & $0.839\pm0.101$ & $0.813\pm0.165$ & $0.950\pm0.100$ & $0.859\pm0.085$ \\
LSTM & $0.811\pm0.065$ & $0.733\pm0.084$ & $1.000\pm0.000$ & $0.843\pm0.058$ \\
\hline
\multicolumn{5}{c}{\textit{3D Fusion}} \\
GRU  & $0.811\pm0.065$ & $0.782\pm0.120$ & $0.900\pm0.085$ & $0.837\pm0.054$ \\
LSTM & $0.846\pm0.145$ & $0.825\pm0.140$ & $0.900\pm0.120$ & $0.861\pm0.134$ \\
\hline
\end{tabular}%
}
\end{table}
\subsection{Ablation Study: 2D vs 3D Feature Comparison}

To validate the effectiveness of 3D representations, an ablation study is conducted comparing against conventional 2D features. The optimal hyperparameters and architectural configuration for 2D feature-based models are shown in Table \ref{tab:2d_hyperparameters}.

\begin{table}[h]
\centering
\caption{Optimal hyperparameters for 2D feature-based RNN architectures.}
\label{tab:2d_hyperparameters}
\begin{tabular}{l|cc|cc}
\hline
\textbf{Parameter} & \multicolumn{2}{c|}{\textbf{2D Facial}} & \multicolumn{2}{c}{\textbf{2D Head Pose}} \\
 & \textbf{GRU} & \textbf{LSTM} & \textbf{GRU} & \textbf{LSTM} \\
\hline
Hidden Size & 64 & 64 & 64 & 64 \\
Num Layers & 2 & 2 & 2 & 1 \\
Dropout & 0.1 & 0.2 & 0.2 & 0.3 \\
Learning Rate & 0.0001 & 0.0001 & 0.001 & 0.001 \\
Batch Size & 16 & 8 & 32 & 16 \\
Bidirectional & No & Yes & No & No \\
\hline
\end{tabular}
\end{table}
\begin{figure*}[h!]
    \centering
    \begin{minipage}{0.48\linewidth}
        \centering
        \includegraphics[width=\linewidth]{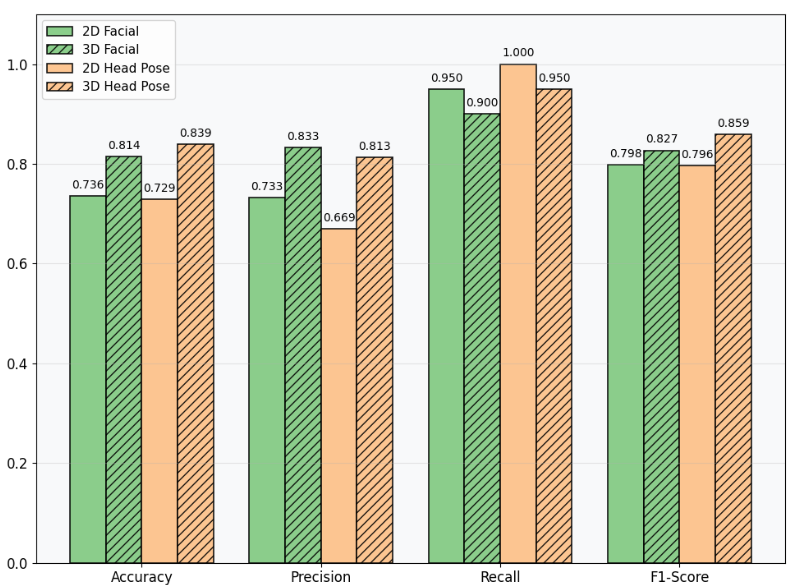}
        \caption*{(a)}
    \end{minipage}
    \hfill
    \begin{minipage}{0.48\linewidth}
        \centering
        \includegraphics[width=\linewidth]{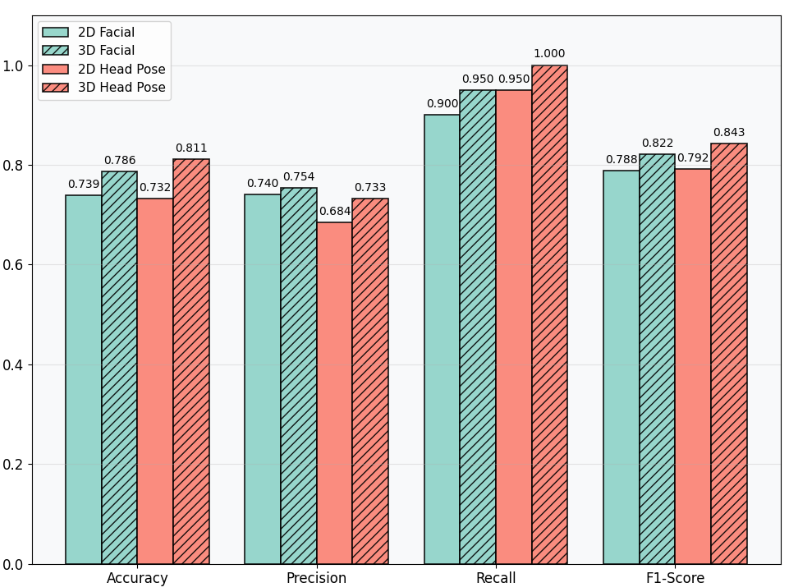}
        \caption*{(b)}
    \end{minipage}
    \caption{Comprehensive performance comparison between 2D and 3D features for (a) GRU and (b) LSTM architectures across facial and head pose features for all evaluation metrics.}
    \label{fig:comprehensive_comparison}
\end{figure*}

The comparative analysis between 2D and 3D feature representations demonstrates substantial performance improvements with 3D temporal modeling across both GRU and LSTM architectures, as shown in Figure \ref{fig:comprehensive_comparison}. Specifically, 3D head pose features outperformed 2D head pose by 10.7\% accuracy, while 3D facial features exceeded 2D facial features by 7.5\% accuracy. These consistent improvements over conventional 2D approaches validate that capturing translational movement components (Tx, Ty, Tz) alongside rotational components and extracting pose-independent facial expressions provides enhanced discriminative information for ASD classification. The 3D head pose representation provides significantly richer behavioral markers than conventional Euler angles alone, while DECA expression parameters capture more nuanced facial dynamics compared to conventional action unit intensities used in prior 2D methods. These findings confirm that 3D temporal modeling successfully captures sequential behavioral patterns characteristic of ASD versus TD children, establishing a foundation for objective, automated screening tools that address current diagnostic limitations in ASD identification for school-age populations.

\subsection{Deployment Considerations}
The proposed framework targets educational screening contexts, such as e-learning platforms, to identify children who may benefit from professional evaluation rather than to provide clinical diagnosis. All data collection followed institutional review board approval with parental consent and participant assent. As a screening tool, false positives may prompt unnecessary referrals while false negatives risk delaying intervention; however, flagged children proceed to specialist assessment, mitigating the impact of misclassification compared to diagnostic errors. Both 3D head pose and facial expression features further support responsible deployment by encoding behavioral patterns as abstract numerical representations without retaining identifiable facial information.
\section{Conclusion}
This study demonstrated the effectiveness of 3D temporal analysis for objective ASD screening during attention tasks. The comprehensive analysis of both rotational and translational components through the DECA framework successfully captured spatial displacement patterns and repetitive movements characteristic of ASD, with 3D head pose features achieving 83.9\% accuracy and 3D facial expression features reaching 81.4\% accuracy, substantially outperforming conventional 2D approaches by 7.5-10.7\% accuracy. Furthermore, multimodal fusion of 3D head pose and facial expression features with PCA-based dimensionality reduction achieved the highest accuracy of 84.6\%, demonstrating that combining complementary behavioral modalities enhances classification performance.
While the current investigation is limited by a relatively small sample size of 39 participants and gender imbalance in the ASD group (17 male, 2 female), which may affect generalizability to broader populations, the promising results establish a foundation for enhanced ASD screening methodologies.
 Future research will focus on expanding the dataset to include larger and more diverse populations while investigating advanced fusion techniques, including late fusion and attention-based mechanisms, to further improve classification performance. The development of such objective, automated screening tools has significant potential to address current clinical limitations in ASD identification, particularly for school-age populations where early intervention can substantially impact developmental outcomes.
\section*{Ethics and Data Availability}
All data collection in this study was conducted with institutional review board approval (Qatar Biomedical Research Institute, Application No. 2017-13), parental consent, and participant assent, ensuring ethical standards are maintained in line with the 1964 Helsinki Declaration. Due to privacy concerns involving minor participants, the raw video dataset is not publicly available. However, extracted features may be provided upon reasonable request, subject to appropriate ethical approvals and data use agreements.
\section*{Acknowledgment}
The research reported in this publication was supported by the Qatar Research Development and Innovation Council (ARG01-0508-230097). The content is solely the responsibility of the authors and does not necessarily represent the official views of the Qatar Research Development and Innovation Council.


{\small
\bibliographystyle{ieee}
\bibliography{egbib}
}

\end{document}